\documentclass[lettersize,journal]{IEEEtran}
\usepackage{amsmath,amsfonts}
\usepackage{algorithmic}
\usepackage{algorithm}
\usepackage{array}
\usepackage[caption=false,font=normalsize,labelfont=sf,textfont=sf]{subfig}
\usepackage{textcomp}
\usepackage{stfloats}
\usepackage{url}
\usepackage{verbatim}
\usepackage{graphicx}
\usepackage{cite}
\usepackage{multirow}
\usepackage{acronym}
\usepackage{booktabs}
\usepackage{pifont}
\usepackage{bm}
\usepackage{tabularx}
\usepackage{threeparttable} 
\usepackage{makecell}

\usepackage[commandnameprefix=always,final]{changes}

\newacro{CNNs}{Convolutional Neural Networks}
\newacro{GANs}{Generative Adversarial Networks}
\newacro{AC-Attention}{Attentive Contextual Attention}
\newacro{CA}{Contextual Attention}

\begin{document}

\title{Attentive Contextual Attention for Cloud Removal}

\author{Wenli Huang, Ye Deng, Yang Wu, and Jinjun Wang 
\thanks{Manuscript was received on August 23, 2024, date of current version October 1, 2024. This work was supported by 
	the Fundamental Research Funds for the Central Universities (Grant No. JBK2103012), the Sichuan Science Foundation project (Grants No. 2024ZDZX0002 and 2024NSFTD0054), and the Ningbo University of Technology Research Fund Project (Grant No. 2040011540031). (Corresponding author: Ye Deng.)}
\thanks{
	Wenli Huang is with the School of Electronic and Information Engineering, Ningbo University of Technology, No. 201, Fenghua Road, Jiangbei District, Ningbo, Zhejiang, China (e-mail: huangwenwenlili@stu.xjtu.edu.cn).
	
	Ye Deng is with the Engineering Research Center of Intelligent Finance, Ministry of Education, School of Computing and Artificial Intelligence, Southwestern University of Finance and Economics, No. 555 Liutai Avenue, Wenjiang District, Chengdu, Sichuan, 611130, China (e-mail: dengye@swufe.edu.cn).
	
	Yang Wu and Jinjun Wang are with the Institute of Artificial Intelligence and Robotics, Xi'an Jiaotong University, Xi'an, Shaanxi 710049, China
	(e-mail: wuyang\_cc@stu.xjtu.edu.cn; jinjun@mail.xjtu.edu.cn).}}

\markboth{IEEE TRANSACTIONS ON GEOSCIENCE AND REMOTE SENSING,~Vol.~14, No.~1, Oct.~2024}%
{Shell \MakeLowercase{\textit{et al.}}: A Sample Article Using IEEEtran.cls for IEEE Journals}

\IEEEpubid{0000--0000/00\$00.00~\copyright~2024 IEEE}

\maketitle

\begin{abstract}

Cloud cover can significantly hinder the use of remote sensing images for Earth observation, prompting urgent advancements in cloud removal technology. Recently, deep learning strategies, especially convolutional neural networks (CNNs) with attention mechanisms, have shown strong potential in restoring cloud-obscured areas. These methods utilize convolution to extract intricate local features and attention mechanisms to gather long-range information, improving the overall comprehension of the scene.
However, a common drawback of these approaches is that the resulting images often suffer from blurriness, artifacts, and inconsistencies. This is partly because attention mechanisms apply weights to all features based on generalized similarity scores, which can inadvertently introduce noise and irrelevant details from cloud-covered areas.
To overcome this limitation and better capture relevant distant context, we introduce a novel approach named Attentive Contextual Attention (AC-Attention). This method enhances conventional attention mechanisms by dynamically learning data-driven attentive selection scores, enabling it to filter out noise and irrelevant features effectively. By integrating the AC-Attention module into the DSen2-CR cloud removal framework, we significantly improve the model's ability to capture essential distant information, leading to more effective cloud removal.
Our extensive evaluation of various datasets shows that our method outperforms existing ones regarding image reconstruction quality. Additionally, we conducted ablation studies by integrating AC-Attention into multiple existing methods and widely used network architectures. These studies demonstrate the effectiveness and adaptability of AC-Attention and reveal its ability to focus on relevant features, thereby improving the overall performance of the networks. The code is available at \url{https://github.com/huangwenwenlili/ACA-CRNet}.

\end{abstract}


\section{Introduction}


\IEEEPARstart{R}{emote} sensing technology has advanced, improving the quality and expanding the applications of remote sensing images. However, cloud cover remains a significant obstacle, obscuring Earth's surface and rendering approximately 67\% of remote sensing data unusable~\cite{king2013spatial}. This data loss impacts further analysis and applications~\cite{ji2020simultaneous}, making cloud removal a critical area of research that has garnered considerable attention in recent studies.

Cloud removal is a critical aspect of addressing missing data in remote sensing and involves various approaches, including spatial-based~\cite{meraner2020cloud}, spectral-based~\cite{meraner2020cloud}, temporal-based~\cite{ebel2022sen12ms}, and hybrid~\cite{ebel2020multisensor} methods. While these techniques can effectively produce cloud-free images, they have several limitations. Many approaches rely heavily on additional data, such as spectral, temporal, and multisensor inputs, which can complicate recovery. Additionally, these methods often require significant computational resources and may struggle to handle cloud cover in complex scenarios. Their effectiveness is generally confined to specific conditions, such as areas with thin clouds or minimal cloud coverage.

\IEEEpubidadjcol

Recent advancements in cloud removal techniques have been driven by deep learning methods that leverage \chreplaced{convolutional}{convolution} operators to extract complex features for effective cloud removal. For instance, the DSen2-CR method~\cite{meraner2020cloud} employs a residual \chreplaced{convolutional}{convolution} network to remove clouds successfully. Similarly, frameworks like Pix2Pix~\cite{isola2017image} and conditional-GAN models~\cite{enomoto2017filmy,bermudez2019synthesis,sarukkai1912cloud} use \chreplaced{convolutional}{convolution} networks to learn data distributions and generate cloud-free images.
However, \chreplaced{convolutional}{convolution} operators have inherent limitations due to their local receptive fields and fixed weights during inference, which can hinder performance. These constraints often cause convolutional neural networks (CNNs) to struggle with diverse and irregular cloud cover, leading to blurry textures, artifacts, and inconsistencies in the resulting images.

\begin{figure*}[!t]
	\centering
	\includegraphics[width=7in]{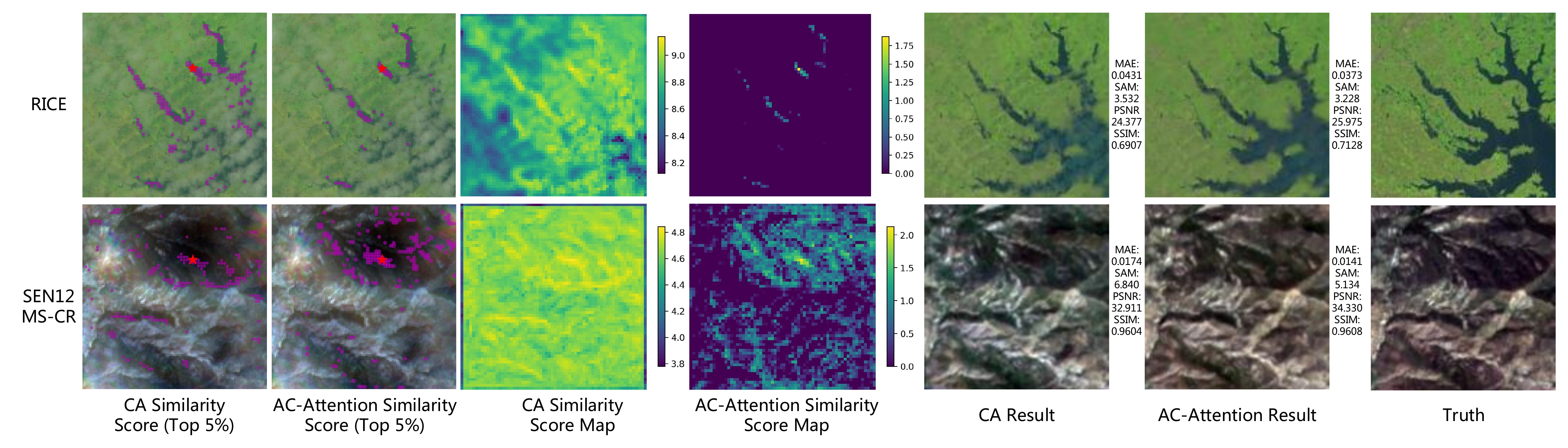}
	\caption{A comparison between the Contextual Attention (CA) and our proposed Attentive Contextual Attention (AC-Attention). 
		The first two columns show a selected query feature, marked by red stars, along with features within the top 5\% similarity to the query feature, marked by magenta pentagons. The following two columns display the similarity scores for this query feature. The next two columns present the cloud removal results.}
	\label{fig:Intro}
\end{figure*}

Researchers have also introduced attention mechanisms to enhance cloud removal techniques by capturing long-range dependencies. For example, UnCRtainTS~\cite{ebel2023uncrtaints} uses self-attention to integrate temporal features across multiple time steps, improving cloud removal effectiveness. SpA GAN~\cite{pan2020cloud} applies a local-to-global spatial attention mechanism focused on cloud areas, which enhances the quality of the resulting cloud-free images. Additionally, \ac{CA}~\cite{2018Generative} assigns weights to global feature patches based on similarity scores, helping generate new image patches.
While attention mechanisms improve the consistency of restored images by expanding the receptive fields, they often rely on dense similarity matrices computed with the softmax function to weigh all global features. This approach can unintentionally introduce irrelevant or invalid features with low scores, reducing the effectiveness of the global receptive fields. Fig.~\ref{fig:Intro} illustrates this problem, showing how similarity scores assigned by the CA mechanism are distributed across the entire spatial region, with many high-scoring points corresponding to invalid features.

In this study, we introduce a novel attention mechanism called \ac{AC-Attention}, designed to identify informative distant patterns for cloud removal in images more effectively. \ac{AC-Attention} employs two trainable modules to calculate the weights and biases for a linear transformation, which helps filter out unhelpful or inaccurate feature connections while strengthening the relationships between relevant features within the similarity score matrix.
Compared to the \ac{CA} mechanism, our \ac{AC-Attention} shows superior performance in cloud removal tasks, as demonstrated in Fig.~\ref{fig:Intro}. Specifically, \ac{AC-Attention} adjusts the similarity scores more effectively, excluding invalid features. This is reflected in both the similarity score map and the images of the top 5\% points, where \ac{AC-Attention} assigns higher scores to the most relevant features. As a result, the cloud removal results using our attention mechanism exhibit improved visual quality and achieve higher quantitative metrics.



We integrated our AC-Attention mechanism into the existing DSen2-CR cloud removal network~\cite{meraner2020cloud}, creating an enhanced model called ACA-CRNet. This 18-layer network includes both residual and AC-Attention blocks. Our evaluation of the RGB datasets RICE1, RICE2~\cite{lin2019remote}, and the multispectral SEN12MS dataset~\cite{meraner2020cloud} demonstrates that ACA-CRNet outperforms existing cloud removal methods. Additionally, ablation studies integrating AC-Attention into several existing methods and widely used network architectures confirm its effectiveness and adaptability, especially in capturing important long-range dependencies.

The contributions of this paper are summarized as follows:

\begin{itemize}
	
	\item We introduce Attentive Contextual Attention (AC-Attention), designed to capture informative long-distance features and eliminate irrelevant or unimportant patterns.
	\item We integrate \ac{AC-Attention} into our ACA-CRNet to improve its cloud removal capabilities and demonstrate its effectiveness.
	\item We evaluate our approach on three diverse datasets, yielding both visual and quantitative results that outperform existing cloud removal methods. Additionally, ablation studies confirm that AC-Attention effectively attends to informative long-distance contexts, highlighting its adaptability and effectiveness.
	
\end{itemize}

The remainder of this paper is organized as follows. Section II reviews the literature on traditional and deep learning-based cloud removal methods. Section III details the architecture of the proposed AC-Attention and the cloud removal network design. Section IV describes the experimental setup, presents the results, and discusses the findings. Finally, Section V summarizes the conclusions.

\IEEEpubidadjcol

\section{Related Works}

In recent years, cloud removal has emerged as a vital technique for data recovery, gaining significant attention in research and practical applications. It is crucial in remote sensing and aerial imaging, improving observations of Earth's surface. This technique encompasses traditional and deep learning approaches.

\subsection{Traditional Cloud Removal Methods}

Traditional cloud removal techniques can be classified into four categories based on the type of information: spatial-based, spectral-based, temporal-based, and hybrid methods.

Spatial-based methods reconstruct cloud-covered areas using texture information from surrounding cloud-free regions within the spatial dimension without needing extra data. These methods include interpolation techniques~\cite{yu2011kriging}, variational approaches~\cite{he2019remote}, and exemplar-based strategies~\cite{he2014image}. These approaches assume that cloud-free and cloud-covered areas have similar statistical properties. They focus on the relationships between local and nonlocal features in the spatial dimension to remove clouds. While these methods are effective for small or thin clouds with simple textures, often struggle with larger, denser, and more complex cloud formations.

To overcome the limitations of spatial-based methods, researchers have developed spectral-based approaches that restore missing data in larger areas by leveraging correlations across different spectral bands. For instance, Wang et al.\cite{wang2006new}, Rakwatin et al.\cite{rakwatin2008restoration}, and Gladkova et al.\cite{gladkova2011quantitative} use the relationships between spectral bands to effectively reconstruct missing data, particularly in band 6 of Aqua MODIS datasets. Meraner et al.\cite{meraner2020cloud} combine multi-spectral optical images with synthetic aperture radar (SAR) images, which are less affected by clouds, to reduce the impact of cloud cover. Xu et al.\cite{xu2019thin} propose a model based on noise-adjusted principal components that integrate spatial and multi-spectral information to produce cloud-free images. Additionally, Teodoro et al.\cite{teodoro2020block} develop a Gaussian mixture model using 3-D patches for hyperspectral image completion. These methods effectively use the strong correlations between different bands in multi-spectral and hyperspectral imagery, making them valuable for cloud removal. However, they are primarily effective for removing thin clouds and may struggle with dense cloud cover that obscures data across all spectral bands.


Some researchers have developed temporal-based methods to address the challenge of cloud removal in images. These methods use data collected at different times to fill in cloudy areas in the same region. For instance, Zeng et al.\cite{zeng2014reconstructing} replaced missing areas with cloud-free content from other periods. Zhu et al.\cite{zhu2011changing} created a temporal filter to restore missing data. Wang et al.\cite{wang2016removing} proposed a temporal low-rank matrix approach that maintains the smoothness and robustness of temporal correlations during cloud removal. Li et al.\cite{li2014recovering} applied a multi-temporal dictionary learning method based on KSVD and Bayesian techniques to recover images affected by dense clouds and shadows. Cao et al.~\cite{cao2020thick} introduced an autoregression-based cloud removal method for Landsat time-series data. These temporal methods effectively use time-based information to remove large areas of thick clouds in simpler scenarios. However, their success depends on having high-quality reference images taken under clear sky conditions, and their performance may be limited in cloud-covered areas.


To achieve better results, researchers have developed hybrid methods that combine information from different domains to remove clouds from remote sensing images. For example, Cheng et al.\cite{cheng2014cloud} used Markov random fields to identify similar pixels by considering spatial and temporal characteristics. Benabdelkader et al.\cite{benabdelkader2008contextual} applied a contextual post-reconstruction approach to capture correlations in the spatial and spectral domains for cloud removal. Li et al.\cite{li2015sparse} used sparse representation techniques to jointly reconstruct missing information by utilizing either spectral-temporal or spatio-temporal data in remote sensing images. Chen et al.\cite{chen2016spatially} developed a spatially and temporally weighted regression method to generate cloud-free Landsat images. Liu et al.\cite{liu2017spatiotemporal} combined temporal and spatial information to reconstruct cloud-covered regions in FengYun-2F land surface temperature data. Ji et al.\cite{ji2018nonlocal} proposed a nonconvex low-rank tensor method to capture global correlations across spatial, spectral, and temporal domains for reconstructing missing information. These methods leverage data from multiple domains (spatial, spectral, or temporal) to improve cloud removal performance. However, they can sometimes encounter inconsistencies between different data domains, leading to artifacts in the restored areas.


\subsection{Deep Learning-based Cloud Removal Methods}


Deep learning methods have recently made significant progress in processing remote sensing images by automatically capturing complex data relationships. Various deep learning architectures, such as convolutional neural networks (CNNs), generative adversarial networks (GANs), and attention-based networks, have been successfully used for cloud removal tasks.

Convolutional neural networks (CNNs) have effectively addressed missing information caused by cloud cover in remote sensing images. For instance, Malek et al.\cite{malek2017reconstructing} used an autoencoder network to explore the relationships between cloud-covered and cloud-free images at pixel and patch levels. Zhang et al.\cite{zhang2018missing} enhanced CNNs by incorporating spatial, spectral, and temporal data to address deadlines, scan line corrector-off problems, and thick cloud removal. Wu et al.\cite{wu2019reconstructing} employed a multiscale encoder-decoder CNN to reconstruct land surface temperature. Li et al.\cite{li2019thin} developed a deep residual network to remove thin clouds from Landsat-8 images. Chen et al.\cite{chen2019thick} combined content, texture, and spectral information to remove thick and thin clouds. Zhang et al.\cite{zhang2020thick} introduced a spatio-temporal network with a global-local loss to handle thick clouds over large areas. Meraner et al.\cite{meraner2020cloud} created a residual neural network for cloud removal in Sentinel-2 images by integrating data from Sentinel-1 synthetic aperture radar (SAR). Ebel et al.\cite{ebel2022sen12ms} used a 3-D CNN to process cloud data across multiple modalities and periods. Czerkawski et al.\cite{czerkawski2022deep} applied the deep image prior method to cloud removal. Xu et al.\cite{xu2023high} introduced Align-CR, which uses low-resolution SAR images to guide cloud removal in high-resolution optical images. These CNN-based methods outperform traditional techniques by optimizing parameters and improving cloud removal performance.

Moreover, Generative Adversarial Networks (GANs) have effectively generated high-quality images with improved clarity and sharpness. Researchers have explored using GANs to reconstruct cloud-free images by learning the underlying data distributions. In this approach, a generator network creates cloud-free images, while a discriminator network is trained to differentiate between generated and real images.
For example, Enomoto et al.\cite{enomoto2017filmy} extended conditional GANs to remove clouds and detect cloud regions using near-infrared and synthetic cloudy images and also struggled with real cloudy images. Singh et al.\cite{singh2018cloud} proposed a cloud-GAN model to remove thin or filmy clouds with cycle consistency loss. Grohnfeldt et al.\cite{grohnfeldt2018conditional} used conditional GANs to combine Sentinel-1 SAR and Sentinel-2 optical images, generating clear cloud-free optical data. Shibata et al.\cite{shibata2018restoration} leveraged reconstruction and adversarial losses to remove cloud occlusions. Bermudez et al.\cite{bermudez2018sar,bermudez2019synthesis} employed conditional GANs with auxiliary SAR data to generate cloud-free optical images, effectively handling thin and thick clouds. Gao et al.\cite{gao2020cloud} proposed a simulation-fusion GAN that translated SAR images into simulated optical images for cloud-free reconstruction. Sarukkai et al.\cite{sarukkai1912cloud} introduced STGAN, which used spatio-temporal data to produce detailed cloud-free images.
However, training GANs can be challenging, often leading to difficulties in convergence and sometimes resulting in unstable predictions.

The deep learning methods discussed above have greatly improved traditional approaches but still have some limitations. These issues stem from the local receptive fields and fixed weights in convolutional operations, which limit the ability of CNNs and GANs to capture broader relationships. As a result, cloud removal can sometimes lead to blurriness, artifacts, and inconsistent results.
To solve these problems, researchers have added attention mechanisms to capture global relationships better and improve the accuracy of cloud-free images. For example, Pan~\cite{pan2020cloud} used a local-to-global spatial attention mechanism to improve image quality. Xu et al.\cite{xu2022glf} introduced a global-local fusion-based cloud removal method (GLF-CR), which includes a shifted-window attention layer for global consistency and dynamic filters for removing clouds based on reliable local texture information. Huang et al.\cite{huang2022ctgan} proposed the CTGAN model, which uses an attention mechanism to extract essential features for cloud removal. Evel et al.\cite{ebel2023uncrtaints} developed the UnCRtainTS model, which applies temporal attention to combine feature maps for cloud removal and uncertainty prediction. Zou et al.\cite{zou2023pmaa} designed the progressive multi-scale attention autoencoder (PMAA) to capture long-range relationships across multiple scales, resulting in detailed, cloud-free images.

While the methods mentioned above use attention mechanisms to capture long-range information and improve cloud removal in deep learning networks, they lack a theoretical analysis of the attention mechanism itself. These methods mainly focus on practical applications and are not explicitly designed for cloud removal tasks. To address this gap, we conducted a study to understand the principles of attention and identified some limitations. Specifically, the attention mechanism uses the softmax function to calculate similarity scores for all global features, which means even irrelevant features can affect global relationships. This can result in artifacts and inconsistencies in the final cloud-free image. To overcome these problems, we propose a new approach called Attentive Contextual Attention (AC-Attention). This novel attention mechanism is designed to capture informative long-range dependencies and is integrated into a residual cloud removal network, making it more effective at handling thick, opaque, and large-scale clouds.

\section{Methodology}

We developed a network called ACA-CRNet, which includes our proposed Attentive Contextual Attention (AC-Attention) mechanism. In this section, we first explain the AC-Attention mechanism and then describe the overall architecture of the ACA-CRNet for cloud removal.


\subsection{Attentive Contextual Attention}
\subsubsection{Vanilla Attention}


\begin{figure*}[!t]
	\centering
	\includegraphics[width=6.5in]{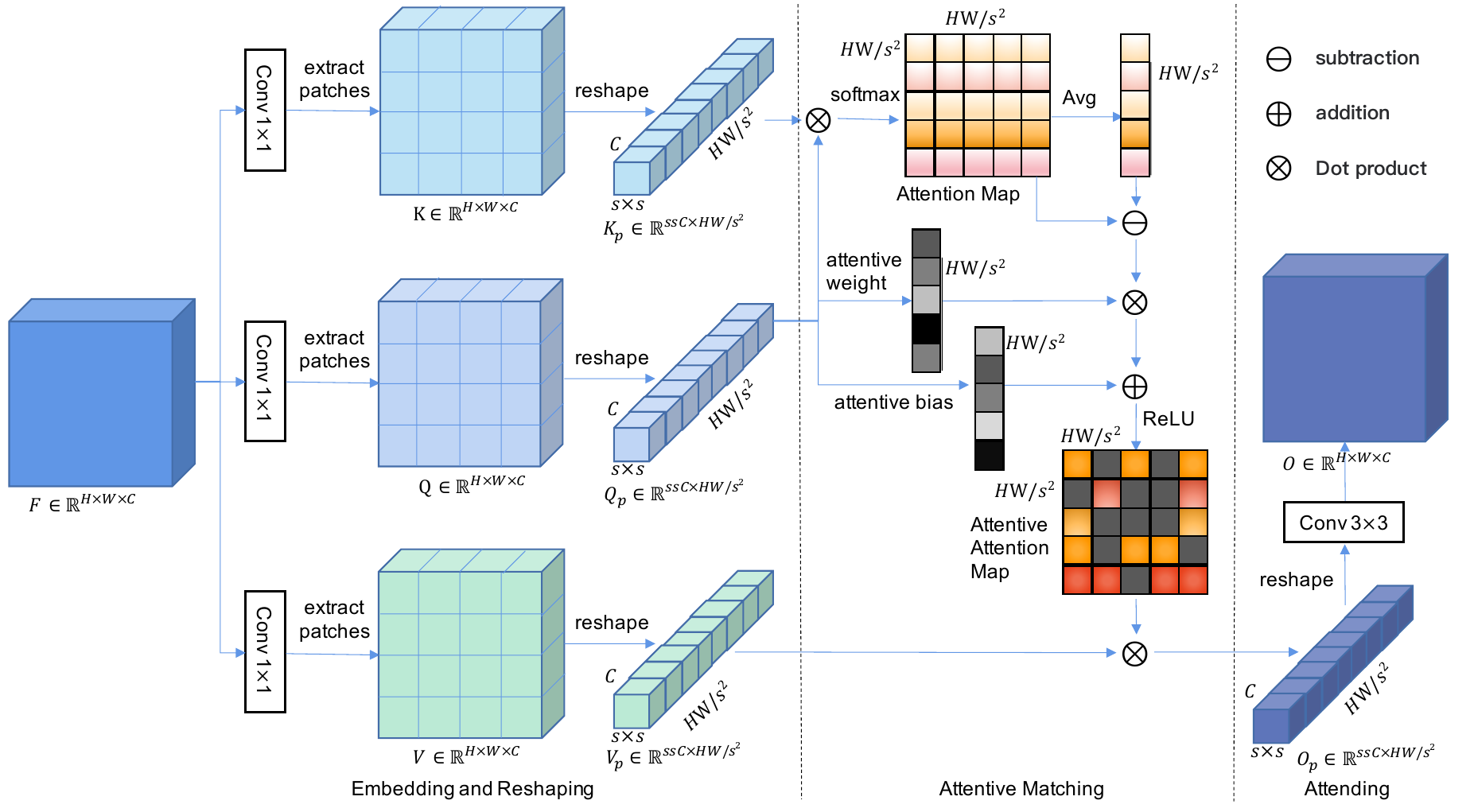}
	\caption{The architecture of Attentive Contextual Attention (AC-Attention), consisting of three steps: embedding and reshaping, attentive matching, and attending. }
	\label{fig:aca-architecture}
\end{figure*}


We begin by reviewing the vanilla attention mechanism introduced by Vaswani et al.~\cite{vaswani2017attention}. Next, we critically analyze its limitations when applied to cloud removal tasks.

We assume that the input feature $F$ is represented as a matrix $F \in\mathbb{R}^{H \times W \times C}$, where $N=H \times W$ is the resolution. The vanilla attention mechanism applies three transformation operators to $F$ to obtain the embedding features: query $Q$, key $K$, and value $V$. These embeddings are represented as matrices: $Q=[q_1,q_2,...,q_N]^T \in\mathbb{R}^{N \times C}$, $K=[k_1,k_2,...,k_N]^T \in\mathbb{R}^{N \times C}$, and $V=[v_1,v_2,...,v_N]^T \in\mathbb{R}^{N \times C}$. This embedding process can be mathematically described as follows:
\begin{equation}
\begin{aligned}
&Q=FW_q, K=FW_k, V=FW_v.
\end{aligned}
\label{eq:embed}
\end{equation}
 Where $W_q, W_k$, and $W_v$ are three trainable $ 1 \times 1$ convolutional filters that transform the input feature $F$ into the embedding space. 
 
 Next, the vanilla attention mechanism calculates a similarity matrix $S\in\mathbb{R}^{N \times N}$ by matching all query and key pairs. This involves a dot-product operation followed by softmax normalization~\cite{vaswani2017attention}. The calculation can be represented as:
 
 \begin{equation}
 \begin{aligned}
 &S=Softmax\left( {QK^T} / {\sqrt{C}} \right),\\
 &s_{i,j}=\frac{exp( { q_i{k_j}^T } / {\sqrt{C}} )}{\sum_{l=1}^N{exp( { q_i{k_l}^T } / {\sqrt{C}} ) } }.
 \end{aligned}
 \label{eq:sall} 
  \end{equation}

 
 

 Finally, the attention mechanism applies the similarity score matrix $S$ to the value $V$, producing the attention output $O=[o_1,o_2,...,o_N]^T \in\mathbb{R}^{N \times C}$. This output is a weighted combination of the values, where the weights are determined by the similarity scores. Mathematically, this is expressed as:
  
 \begin{equation}
 \begin{aligned}
 &O=SV,\\
 &o_{i} = \sum_{j=1}^N{s_{i,j}v_j}.
 \end{aligned}
 \label{eq:attend} 
 \end{equation}

The attention mechanism we discussed calculates the similarity score $s_{i,j}$ between a query and keys in Equation~\ref{eq:sall}. It considers all key features $K=[k_1,k_2,...,k_N]^T$ for each query and applies a softmax operation to ensure positive scores. The sum of similarity scores for each query $\sum_{j=1}^N(S_{i,j})$ is 1, meeting the requirements of a probability distribution. However, there is a drawback to this approach. When the similarity score $s_{i,j}$ is calculated between a query $q_i$ and an irrelevant or invalid key $k_j$, the score is very close to 0. Despite this score is small, it still affects the corresponding attention output $o_i$ in Equation~\ref{eq:attend}, which weakens the ability to capture global relationships. This limitation is unsuitable for cloud removal tasks, often leading to artifacts and inconsistencies in the resulting cloud-free images.


\subsubsection{Attentive Contextual Attention (AC-Attention)}



We observed that some attention mechanisms used in existing cloud removal approaches methods~\cite{pan2020cloud, xu2022glf, huang2022ctgan, ebel2023uncrtaints, zou2023pmaa} include irrelevant long-range dependencies, which can cause issues in the final results. To address this, we propose an Attentive Contextual Attention (AC-Attention) mechanism inspired by Contextual Attention (CA)\cite{2018Generative}. Our AC-Attention is designed to model more appropriate global relationships and focus on relevant features for cloud removal. The architecture, illustrated in Fig.\ref{fig:aca-architecture}, involves three steps: embedding and reshaping, attentive matching, and attending. More details will be provided in the following content.

\paragraph{Embedding and Reshaping}


Similar to previous attention mechanisms~\cite{vaswani2017attention,2018Generative}, we first use Equation~\ref{eq:embed} to map the feature maps $F \in\mathbb{R}^{H \times W \times C}$ into three latent embedding features: $Q$, $K$, and $V$. These features are then divided into $N_p= {HW} / {s^2}$ patches, where each patch has dimensions $d= s \times s \times C$. Subsequently, we flatten and reshape these patches into matrices, denoted as $Q_p=[q_{p,1}, q_{p,2}, ..., q_{p,N_p}]^T\in\mathbb{R}^{N_p \times d}$, $K_p=[k_{p,1}, k_{p,2}, ..., k_{p,N_p}]^T\in\mathbb{R}^{N_p \times d}$, and $V_p=[v_{p,1}, v_{p,2}, ..., v_{p,N_p}]^T\in\mathbb{R}^{N_p \times d}$.

\paragraph{Attentive Matching}


We utilize a patch-based matching operation, inspired by the vanilla attention mechanism~\cite{vaswani2017attention}, to calculate the similarity matrix $S_q$. Similar to the process described in Equation~\ref{eq:sall}, we perform a dot-product operation to calculate the relationships between query patches $Q_p$ and key patches $K_p$. Then, we apply the softmax function to obtain the patch-based similarity matrix $S_p\in\mathbb{R}^{N_p \times N_p}$. This computation can be expressed as:

 \begin{equation}
\begin{aligned}
&S_p=Softmax\left( {Q_pK_p^T} / {\sqrt{d}} \right).
\end{aligned}
\label{eq:sall_query} 
\end{equation}

When a value in the patch-based similarity matrix $S_p$ is small or close to 0, it indicates a weak or non-existent correlation between the corresponding dependencies. To address this, we propose a learnable selection mechanism based on the query features $Q$ to exclude irrelevant or invalid feature relationships from $S_p$. This mechanism consists of two modules: a weight module and a bias module. These modules are responsible for learning the weight $W$ and bias $B$, respectively, and use them in a linear transformation function to adjust the similarity matrix. 

Specifically, the learnable module first reduces the query feature channels from $C$ to $C/4$ using a convolution, activates the features with a Rectified Linear Unit (ReLU) function~\cite{nair2010rectified}, and then further reduces the feature channels to 1 using another convolution. This process can be represented as:


\begin{equation}
\begin{aligned}
&W = Conv\left(ReLU\left(Conv(Q)\right)\right),\\
&B = Conv\left(ReLU\left(Conv(Q)\right)\right).
\end{aligned}
\label{eq:sall_weights_bias} 
\end{equation}


Next, we compute the average similarity value for each query in the similarity matrix $S_p$, represented as $S_{p,avg} \in\mathbb{R}^{N_p}$. This average value is obtained by taking the mean of each $N_p$-dimensional row vector in $S_p$. Then, we generate an adjusted similarity matrix $S_{p,ad}$ by subtracting the average similarity value $S_{p,avg}$ from the original matrix $S_p$. Mathematically, this process is expressed as:

 \begin{equation}
\begin{aligned}
&S_{p,avg} = Avg(S_p), \\
&s_{p,avg,i} = \frac{\sum_{j=1}^{N_p}{ s_{p,i,j} }}  {N_p },\\
&S_{p,ad} = S_p-S_{p,avg}.
\end{aligned}
\label{eq:sall_ad} 
\end{equation}

 
 
Finally, we apply the adjusted similarity matrix $S_{p,ad}$ to a linear transformation function using the weight $W$ and bias $B$ to obtain the attentive similarity matrix $S_{att}$. Since this matrix might include negative values, which indicate uncorrelated or noisy relationships, we apply the ReLU function to filter out these negatives. This process is represented as:
 
\begin{equation}
 \begin{aligned} 
 &S_{att} = ReLU\left[  S_{p,ad} * W + B \right].
 \end{aligned}
 \label{eq:sall_attentive} 
 \end{equation}

\paragraph{Attending}




After obtaining the attentive similarity scores $S_{att}$, we compute the patch-based output $O_p$ by summing the value patches $V_p$ with weights determined by $S_{att}$. This calculation can be expressed as follows:

\begin{equation}
\begin{aligned} 
&O_p = S_{att}V_p.\\
\end{aligned}
\label{eq:sall_attending} 
\end{equation}



Next, we reshape the patch-based output $O_p$ to match the dimensions of $H \times W \times C$, ensuring it aligns with the input feature map $F$. Finally, we apply a $3 \times 3$ convolution to the reshaped feature, resulting in the final output $O$ of our \ac{AC-Attention}.



In summary, our \ac{AC-Attention} introduces a learnable selection mechanism during the attentive matching stage. This mechanism uses two learnable modules to determine the weight and bias for a linear transformation function. It allows adaptive adjustments to the similarity matrix based on the query, filtering out irrelevant relationships and enhancing the attention scores for relevant features. This design aims to improve the effectiveness of attention in capturing useful information.

\begin{figure*}[!t]
	\centering
	\includegraphics[width=7in]{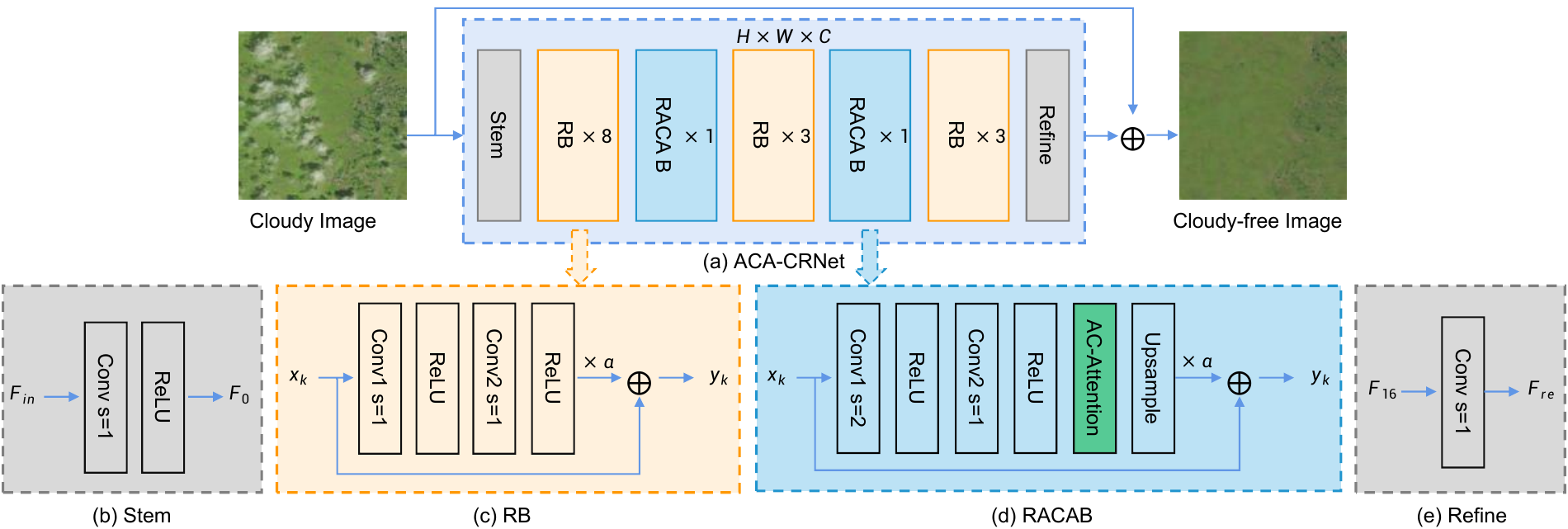}
	\caption{Overview of our proposed ACA-CRNet, including (a) ACA-CRNet architecture, (b) Stem, (c) Residual Block (RB), (d) Residual AC-Attention Block (RACAB), and (e) Refine component. The ACA-CRNet is designed in a residual style, comprising the Stem, RBs, RACABs, and Refine modules.}

	\label{fig:net-architecture}
\end{figure*}

\begin{table}[t]
	\centering
	\caption{Detailed architecture of our ACA-CRNet. }
	\resizebox{.9\columnwidth}{!}  {
		\begin{tabular}{m{4em}<{\centering}m{8em}<{\centering}m{14em}<{\centering}}
			\toprule
			Stage & Operator & Param \\
			\midrule
			Input & -     & H$\times$W$\times$ $C_{in}$ \\
			Stem & Conv, ReLU & $[3 \times 3 \times C_{in} \times C], s=1$ \\
			Stage 1 & [RB] $\times 8$ & Conv1: $[3\times 3 \times C \times C ] , s=1$, Conv2: $[3\times 3 \times C \times C ], s=1 $, $\alpha=0.1$ \\
			Stage 2 & [RACAB] $\times 1$ &Conv1: $[3\times 3 \times C \times C ] , s=2$, Conv2: $[3\times 3 \times C \times C ], s=1 $, $\alpha=0.1$, $upScale=2$  \\
			Stage 3 & [RB] $\times 3$  &Conv1: $[3\times 3 \times C \times C ] , s=1$, Conv2: $[3\times 3 \times C \times C ], s=1 $, $\alpha=0.1$  \\
			Stage 4 & [RACAB] $\times 1$ &  Conv1: $[3\times 3 \times C \times C ] , s=2$, Conv2: $[3\times 3 \times C \times C ], s=1 $, $\alpha=0.1$, $upScale=2$ \\
			Stage 5 & [RB] $\times 3$ & Conv1: $[3\times 3 \times C \times C ] , s=1$, Conv2: $[3\times 3 \times C \times C ], s=1 $, $\alpha=0.1$ \\
			Refine & Conv & [$3 \times 3 \times C \times C_{in}], s=1$ \\
			Output & Concat Input and Refine & H$\times$W$\times$$C_{in}$ \\
			\bottomrule
		\end{tabular}
	}	
	
	\begin{tablenotes}
		\footnotesize
		\item[*] The input has dimensions of height ($H$), width ($W$), and channel ($C_{in}$). The basic channel size is $C$. The notation $[k \times k \times C_{1} \times C_{2}]$ represents a \chreplaced{convolutional}{convolution} filter with $k$ kernel size, $C_{1}$ input channels, and $C_{2}$ output channels. The stride of the convolution is $s$. 
	\end{tablenotes}
	
	\label{tab:net-details}%
\end{table}%

\subsection{Network Architecture}
\chadded{We have designed a cloud removal network named ACA-CRNet, which builds on the residual-style architecture of the DSen2-CR network \cite{meraner2020cloud}. By incorporating the AC-Attention mechanism, ACA-CRNet enhances its ability to capture informative global features, making cloud removal more effective. The architecture of ACA-CRNet is illustrated in Fig.~\ref{fig:net-architecture}. Our design introduces novel Residual AC-Attention Blocks (RACAB), which are integrated into the DSen2-CR framework by replacing two of its Residual Blocks (RBs). The RBs are derived from ResNet \cite{he2016deep}, while the RACABs are specifically designed to leverage the AC-Attention module. In the following sections, we will provide a detailed explanation of the network structure and the composition of these blocks.}


\subsubsection{Architecture}

The DSen2-CR method \cite{meraner2020cloud} has demonstrated the effectiveness of residual connection networks in addressing challenges related to thin cloud correction, cloud-free region reconstruction, predictive stability, and deep model optimization. Building on this approach, we present ACA-CRNet, a novel network architecture that integrates Residual Blocks (RBs) with AC-Attention to enhance cloud removal capabilities. The architecture of ACA-CRNet is illustrated in Figure \ref{fig:net-architecture} (a) and further detailed in Table \ref{tab:net-details}.

ACA-CRNet comprises 18 layers, including 2 \chreplaced{convolutional}{convolution} layers, 14 RB modules, and 2 RACAB modules. The initial layer, referred to as the stem, initiates the feature extraction process by applying a $3 \times 3$ \chreplaced{convolutional}{convolution} filter to transform the input image $F_{in} \in \mathbb{R}^{H \times W \times C_{in}}$ into an initial feature map $F_{0} \in \mathbb{R}^{H \times W \times C}$, where C is configured to align with the baseline channel requirements of the subsequent RBs. Next, a series of 14 RBs systematically extract a rich hierarchy of shallow and deep features. To further improve feature extraction, RACABs are strategically placed at the 10th and 14th layers, allowing the network to capture and use valuable global contextual information, which enhances cloud removal performance. As the features propagate through the network, they undergo refinement and enrichment, culminating in the final refinement layer. Here, a $3 \times 3$ \chreplaced{convolutional}{convolution} filter is applied to transform the deepest feature map $F_{16} \in \mathbb{R}^{H \times W \times C}$ into a refined feature map $F_{re} \in \mathbb{R}^{H \times W \times C_{in}}$, which retains the same spatial and channel dimensions as the input feature map $F_{in}$. Finally, $F_{in}$ is concatenated with the refined feature map $F_{re}$ to produce the final output $F_{out}$.

To effectively train ACA-CRNet, we use the reconstruction loss function, which has proven successful in previous cloud removal studies \cite{meraner2020cloud,ebel2023uncrtaints}. This loss function helps the network improve its cloud removal performance by minimizing the difference between the cloud-free output and the corresponding ground truth image.





\subsubsection{Residual Block}

The Residual Block (RB) is the fundamental module in our ACA-CRNet and is derived from the DSen2-CR network~\cite{meraner2020cloud}. The architecture of RB is illustrated in Fig.~\ref{fig:net-architecture} (c) and can be described as follows:

\begin{equation}
\begin{aligned} 
&x_k' = ReLU( Conv( ReLU( Conv(x_k) ))),\\
&y_k = x_k + \alpha x_k'.
\end{aligned}
\label{eq:rb} 
\end{equation}
Where $x_k$ and $y_k$ represent the input and output features of the $k$-th RB module, respectively. First, $x_k$ is passed through two sets of $3 \times 3$ convolution followed by ReLU activation, resulting in the feature $x_k'$. Next, $x_k'$ is multiplied by a scaling factor $\alpha$ to produce the main branch output $x_k''$, which helps stabilize the training process. Typically, $\alpha$ is set to 0.1. Finally, the scaled output $x_k''$ is added to the input feature $x_k$ to produce the final result $y_k$ of the RB module.

\subsubsection{Residual AC-Attention Block}

To capture global relationships, we incorporated our proposed \ac{AC-Attention} into the RB module, creating the Residual AC-Attention Block (RACAB). The structure of RACAB is illustrated in Fig.~\ref{fig:net-architecture} (d). The formulation of RACAB can be described as follows:
\begin{equation}
\begin{aligned} 
&x_k' = Up (ACA( ReLU( Conv( ReLU( Conv(x_k) ))) ) ),\\
&y_k = x_k + \alpha x_k'.
\end{aligned}
\label{eq:rab} 
\end{equation}
Where the $ACA$ denotes our AC-Attention module, the $Up$ denotes the upsampling operation. To reduce computational complexity, we set the stride of the first $3 \times 3$ convolution to 2, denoted as $s=2$. This setting helps decrease the attention module's computational cost, which increases quadratically with the spatial resolution of the input feature. The AC-Attention is applied after the second ReLU activation. Finally, we use interpolation to upsample the feature back to its original spatial resolution. All other operations and settings remain the same as in the RB module.



\begin{table}[t]
	\centering
	\caption{Publicly available datasets used for cloud removal experiments. }
	\resizebox{.9\columnwidth}{!}  {
		\begin{tabular}{m{6em}<{\centering}m{5em}<{\centering}m{4em}<{\centering}m{4em}<{\centering}m{4em}<{\centering}m{2em}<{\centering}}
			\toprule
			Dataset &Source& Resolution & Train images& Test images & w/ SAR\\
			\midrule
			RICE-I~\cite{lin2019remote} & Google Earth & $<$15m & 450 & 50 & \ding{55} \\
			RICE-II~\cite{lin2019remote} &  Landsat-8 & 30m & 630 & 106 &  \ding{55} \\
			SEN12MS-CR~\cite{meraner2020cloud} & Sentinel-2 & 10m & 101,615 & 7,899 &  \ding{51} \\
			\bottomrule
		\end{tabular}%
	}	
	\label{tab:dataset}%
\end{table}%

\section{Experiment}

This section first describes the datasets, \chreplaced{evaluation}{elvaluation} metrics, and implementation details used in the experiments. Next, we evaluate the performance of our approach against five state-of-the-art cloud removal methods. Finally, we perform ablation studies by integrating AC-Attention into various existing methods and network architectures to evaluate its effectiveness and adaptability.

\subsection{Experiments Details}
\subsubsection{Datasets}

We evaluated our cloud removal approach using three publicly available datasets, as summarized in Table~\ref{tab:dataset}.

The first dataset, RICE-I~\cite{lin2019remote}, consists of 500 pairs of filmy and cloud-free images obtained from Google Earth. The cloud-free images were generated by manually adjusting the cloud layer's visibility.

The second dataset, RICE-II~\cite{lin2019remote}, consists of 736 pairs of images captured by Landsat 8 OLI/TIRS, including cloudy, cloudless, and mask images. The mask images were created using the Landsat Level-1 quality band to identify regions affected by clouds, cloud shadows, and cirrus clouds. The cloudless images were captured at the same location as the corresponding cloud images with a maximum interval of 15 days. This dataset features heavy and complex cloud coverage.

The third dataset, SEN12MS-CR~\cite{meraner2020cloud}, contains approximately 110,000 samples from 169 distinct, non-overlapping regions across various continents and meteorological seasons. Each sample includes a pair of Sentinel-2 images, one cloudy and one cloud-free, along with the corresponding Sentinel-1 synthetic aperture radar (SAR) image. This dataset provides a diverse and comprehensive range of real-world cloud scenarios across different environmental settings.


\subsubsection{Metrics}

Following previous cloud removal studies~\cite{meraner2020cloud, xu2022glf, sarukkai1912cloud, ebel2023uncrtaints}, we used several quantitative metrics to evaluate the performance of our cloud removal approach. These metrics include mean absolute error (MAE) \cite{chai2014root}, peak signal-to-noise ratio (PSNR) \cite{korhonen2012peak}, structural similarity index (SSIM) \cite{wang2004image}, and spectral angle mapper (SAM) \cite{kruse1993spectral}. The mathematical definitions of these metrics are as follows:
\begin{equation}
\begin{aligned}
&MAE\left( X,Y \right) = \frac{1}{ N} \sum_{i=1}^{N}{\|x_{i} - y_{i}  \|_1},\\
&MSE\left( X,Y \right) = \frac{1}{ N} \sum_{i=1}^{N}{\left(x_{i} - y_{i}  \right)^2},\\
&PSNR\left(X,Y \right) = 10 \cdot {log} \left(  \frac{L^2}{ MSE} \right),\\
&SSIM\left( X,Y \right) =  \frac{(2\mu_x\mu_y+C_1)(2\sigma_{x,y}+C_2)}{(\mu_x^2+\mu_y^2+C_1)(\sigma_x^2+\sigma_y^2+C_2)},\\
&SAM\left( X,Y \right) = cos^{-1}\left( \frac{Y^TX}  { \|Y \| \|X \|}   \right).
\end{aligned}
\label{equation:metrics}
\end{equation} 



In defining the evaluation metrics, we use $X$ as the predicted image and $Y$ as the ground truth image. The individual elements of $X$ and $Y$ are denoted as $x_i$ and $y_i$, respectively, with $N$ representing the total number of pixels in the image. The maximum value of the digital image is set as $L=255$. The mean values of $X$ and $Y$ are denoted as $\mu_x$ and $\mu_y$, while the standard deviations are  $\sigma_x$ and $\sigma_y$. The covariance between $X$ and $Y$ is indicated as $\sigma_{x,y}$. To avoid division by zero, two constants are used: $C_1=(K_1L)^2$ and $C_2=(K_2L)^2$, with default values $K_1=0.01$ and $K_2=0.03$.

The MAE and PSNR metrics assess similarity at the pixel level, while SSIM measures spatial structural quality. SAM evaluates spectral fidelity by calculating the angle between vectors. Lower MAE and SAM values indicate better reconstruction, whereas higher PSNR and SSIM values reflect superior reconstruction performance.


\subsubsection{Implementation Details}


We implemented our code using PyTorch, an open-source Python library for machine learning. The training was performed on a single NVIDIA GeForce RTX 3090 GPU with 24 GB of memory, running on Ubuntu 20.04. We used the Adam optimizer with betas (0.9, 0.999) and a $7 \times 10^{-5}$ learning rate. The base channel size ($C$) was set to 256. For the RICE-I and RICE-II datasets, we randomly cropped samples to a size of $128 \times 128$. The models were trained with a batch size of 12 for 300 epochs. For the SEN12MS-CR dataset, the sample size was set to $256 \times 256$, and the models were trained with a batch size of 6 for 30 epochs. Bilinear interpolation was used to upsample feature maps in the Residual AC-Attention Block (RACAB). The code and pre-trained models will be made publicly available on GitHub.

\begin{table}[tbp]
	\centering
	\caption{Quantitative comparison results on three datasets. }   
	\resizebox{.99\columnwidth}{!}  {
		\begin{tabular}{cccccc}
			\toprule
			\textbf{Datasets} &\textbf{Methods} & \textbf{MAE} $\downarrow$ & \textbf{SAM} $\downarrow$ & \textbf{PSNR}$\uparrow$ & \textbf{SSIM}$\uparrow$ \\
			\midrule
			\multirow{6}[1]{*}{\makecell[c]{RICE-I}} & pix2pix~\cite{isola2017image} & 0.032  & 2.370  & 29.70  & 0.893  \\
			&SpA GAN~\cite{pan2020cloud}\tnote{*} & 0.040  & 2.352  & 28.04  & 0.915  \\
			&STGAN~\cite{sarukkai1912cloud} & 0.024  & 1.701  & 32.38  & 0.956  \\
			&DSen2-CR~\cite{meraner2020cloud}\tnote{*} & 0.020  & 1.278  & 33.65  & 0.975  \\
			&UnCRtainTS~\cite{ebel2023uncrtaints} \tnote{*} & 0.022  & 1.462  & 33.25  & 0.972  \\			&Ours  & $\bm{0.014}$   & $\bm{1.078}$   & $\bm{36.66}$   & $\bm{0.979}$   \\
			\midrule
			\multirow{6}[1]{*}{\makecell[c]{RICE-II}} & pix2pix~\cite{isola2017image} & 0.030  & 2.144  & 30.54  & 0.799  \\
			&SpA GAN~\cite{pan2020cloud} & 0.037  & 2.974  & 28.21  & 0.810  \\
			&STGAN~\cite{sarukkai1912cloud} & 0.030  & 2.691  & 29.91  & 0.820  \\
			&DSen2-CR~\cite{meraner2020cloud} & 0.022  & 1.859  & 32.34  & 0.880  \\
			&UnCRtainTS~\cite{ebel2023uncrtaints} & 0.025  & 1.927  & 31.82  & 0.871  \\
			&Ours  & $\bm{0.020}$   & $\bm{1.600}$   & $\bm{33.70}$   & $\bm{0.891}$  \\
			\midrule
			\multirow{7}[1]{*}{\makecell[c]{SEN12\\MS-CR}} & pix2pix~\cite{isola2017image} & 0.031  & 10.784  & 27.60  & 0.864  \\
			&SpA GAN~\cite{pan2020cloud}$^{*}$  & 0.045  & 18.085  & 24.78  & 0.754  \\
			&STGAN~\cite{sarukkai1912cloud} & 0.034  & 12.798  & 26.50  & 0.838  \\
			&DSen2-CR~\cite{meraner2020cloud}$^{*}$ & 0.031  & 9.472  & 27.76  & 0.874  \\
			&GLF-CR~\cite{xu2022glf}$^{*}$  & 0.028  & 8.981  & 28.64  & 0.885  \\
			&UnCRtainTS~\cite{ebel2023uncrtaints} $^{*}$  & 0.027  & 8.320  & 28.90  & 0.880  \\
			&Ours  & $\bm{0.025}$  & $\bm{7.770}$  & $\bm{29.78}$  & $\bm{0.896}$  \\
			\bottomrule
		\end{tabular}%
	}
		\begin{tablenotes}
			\footnotesize
			\item[*] The evaluation metrics for the methods marked with * are sourced from the UnCRtainTS paper~\cite{ebel2023uncrtaints}. We assessed the performance of the other methods using models we trained based on their published code.
			The $\downarrow$ indicates lower is better, while $\uparrow$ indicates higher is better. Bold numbers indicate the best performance achieved in each metric.
		\end{tablenotes}

	\label{tab:performanc-3datasets}%
\end{table}%

\begin{figure*}[!t]
	\centering
	\includegraphics[width=7in]{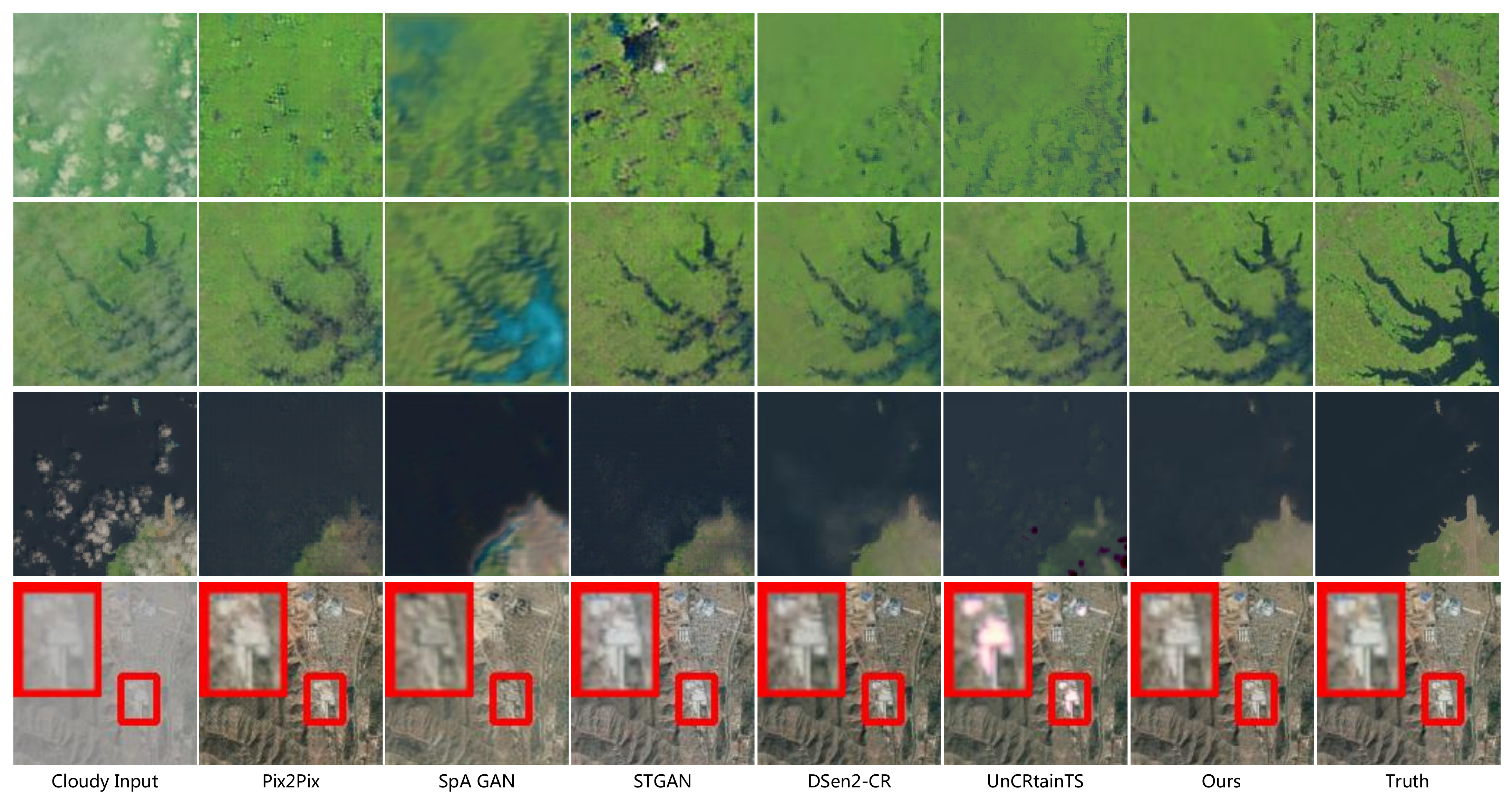}
	\caption{Visualization of cloud removal results on RICE-I and RICE-II datasets. Local details are highlighted in red boxes. Zooming in is recommended for a clearer view.}
	\label{fig:performance-rice}
\end{figure*}

\begin{figure*}[!t]
	\centering
	\includegraphics[width=7in]{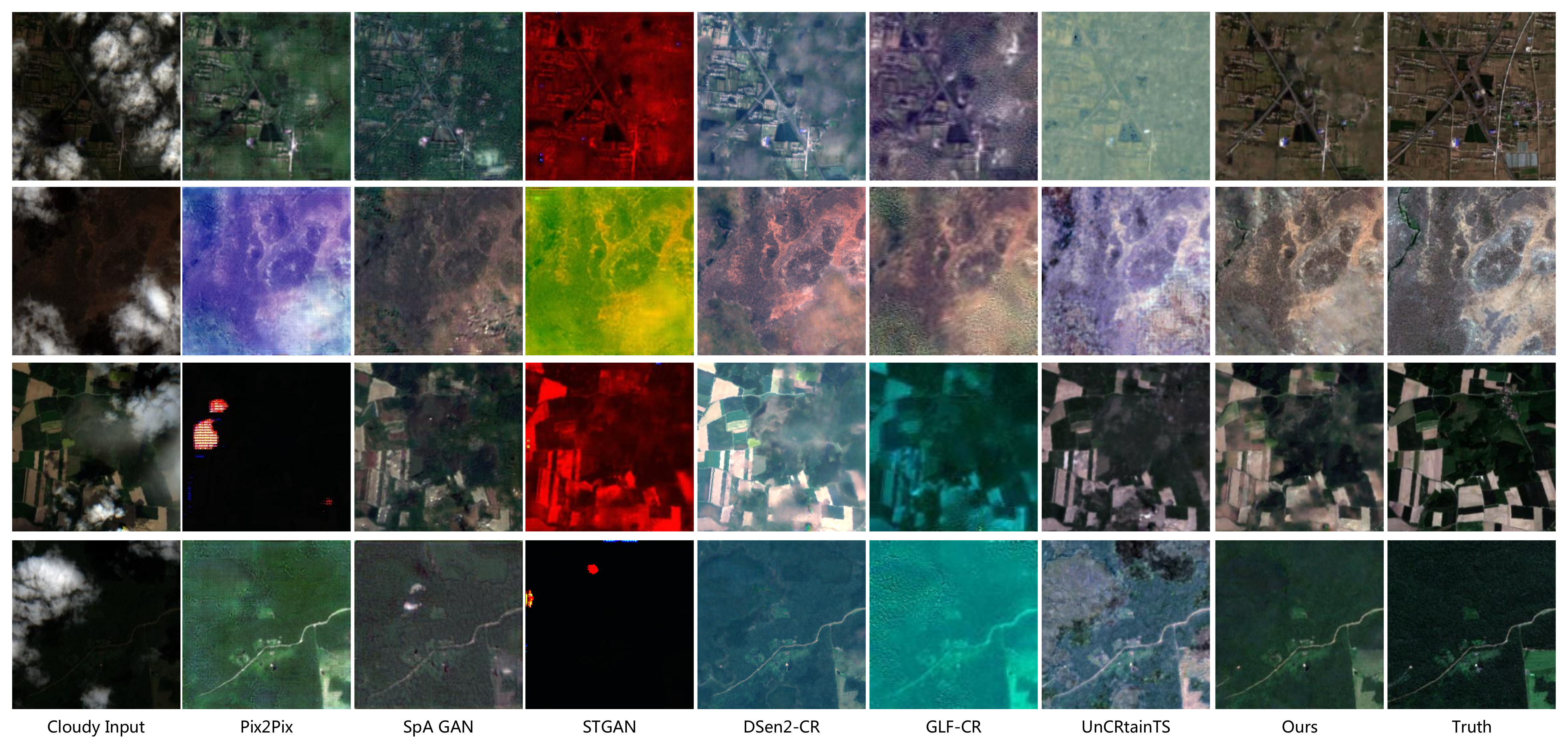}
	\caption{Visualization of cloud removal results for Sentinel-2 satellite data on the SEN12MS-CR dataset. The Sentinel-2 data includes 13 spectral bands, with the visualization images generated using the RGB bands. Zooming in is recommended for a clearer view.}
	\label{fig:performance-sen12mscr}
\end{figure*}

\subsection{Performance of ACA-CRNet}

\subsubsection{Comparison Baselines}

To evaluate the performance of our proposed network, we compared it to six advanced deep-learning models for cloud removal. 

\begin{itemize}
	\item{Pix2Pix~\cite{isola2017image}:} A variant of conditional GANs that employs an encoder-decoder framework for image-to-image translation tasks.
	\item{SpA GAN~\cite{pan2020cloud}:} A GAN-based approach that uses local-to-global spatial attention to identify and reconstruct cloud-free regions.
	\item{STGAN~\cite{sarukkai1912cloud}:} A GAN-based spatio-temporal generator network that tackles cloud removal as a conditional image synthesis problem by utilizing temporal and infrared data.
	\item{DSen2-CR~\cite{meraner2020cloud}:} A residual network that integrates synthetic aperture radar (SAR) data with a cloud-adaptive regularized loss to restore cloudless images.
	\item{GLF-CR~\cite{xu2022glf}:} A network that uses global-local fusion techniques, including SAR-guided blocks, shifted-window attention, and dynamic convolution, to generate reliable textures and consistent structures in cloud-covered areas.
	\item{UnCRtainTS~\cite{ebel2023uncrtaints}:} A model that combines multivariate uncertainty regression with attention-based temporal encoding to restore images obscured by dense clouds.
\end{itemize}

Among the baseline models, Pix2Pix, SpA GAN, and STGAN use GAN architectures to generate cloud-free images. SpA GAN, GLF-CR, and UnCRtainTS incorporate attention mechanisms to capture global information. DSen2-CR uses a deep residual convolutional network to restore cloud-affected regions.
GLF-CR is evaluated only on the SEN12MS-CR dataset because it requires SAR data for fusion. We obtained the SpA GAN, DSen2-CR, GLF-CR, and UnCRtainTS models for the DSen2-CR dataset from the UnCRtainTS paper~\cite{ebel2023uncrtaints}. The other models were trained on our experimental datasets using the publicly available code provided in their respective papers.


\subsubsection{\chreplaced{Quantitative}{Quantiative} Comparison with Baselines}

\chreplaced{Table}{Tables}~\ref{tab:performanc-3datasets} shows the quantitative comparison results for the RICE-I, RICE-II, and SEN12MS-CR datasets. 
The table uses intuitive symbols: $\downarrow$ indicates that lower values are better, while $\uparrow$ indicates that higher values are preferable. The best metric values for each dataset are highlighted in bold, clearly demonstrating the superiority of our approach.

Our ACA-CRNet consistently outperforms all baseline methods across the four evaluated metrics on all three datasets. Specifically, compared to the latest UnCRtainTS methods, ACA-CRNet achieves significant improvements in the MAE and SAM metrics, with average gains of 30.05\% and 21.05\%, respectively, across the datasets. Additionally, our method shows average improvements of 5.95\% in PSNR and 1.58\% in SSIM.

\chadded{Our method's exceptional performance on the RICE-I dataset likely stems from its characteristics. This dataset's simpler scenes and translucent thin clouds retain more usable information, making cloud removal more efficient. These factors, combined with our approach's strengths, lead to even greater performance gains on the RICE-I dataset.}

\subsubsection{Qualitative Comparison with Baselines}

We visually compare the cloud removal results of our method with five baselines on the RICE-I and RICE-II datasets in Fig.~\ref{fig:performance-rice}. 

Our method successfully restores consistent scene details, such as recovering the grass texture in the first row and preserving the sea image in the third row. It also maintains the shape and integrity of objects, as demonstrated by the river in the second row and the island in the third row. Additionally, our method effectively preserves object details when removing translucent thin clouds, as highlighted in the red box in the fourth row. In contrast, the other five methods show color anomalies, artifacts, incomplete structures, and missing objects in their results.

We also qualitatively compare the cloud removal results with six baselines on the SEN12MS-CR dataset in Fig.~\ref{fig:performance-sen12mscr}. Our method accurately restores scene colors, closely matching the ground truth. In contrast, the other six methods introduce red, yellow, or purple color distortions in the result images. Additionally, our method effectively restores edges, structures, and details, such as roads in the third row and buildings in the fourth row.

Overall, our method outperforms the six advanced baselines based on quantitative and qualitative analysis.

\subsection{Ablation Studies}

\begin{table*}[tbp]
  \centering
  \caption{\chadded{Effectiveness of our AC-Attention on three network architectures}}
  \resizebox{.9\textwidth}{!}  {
    \begin{tabular}{cc|cccc|cccc|cccc}
    \toprule
    \multicolumn{2}{c|}{\textbf{Networks}} & \multicolumn{4}{c|}{\textbf{ACA-CRNet}} & \multicolumn{4}{c|}{\textbf{ACA-EDNet}} & \multicolumn{4}{c}{\textbf{ACA-Unet}} \\
    \midrule
    Datasets & Methods & MAE$\downarrow$  & SAM$\downarrow$ & PSNR$\uparrow$  & SSIM$\uparrow$  & MAE$\downarrow$ & SAM$\downarrow$ & PSNR$\uparrow$  & SSIM$\uparrow$  & MAE$\downarrow$ & SAM$\downarrow$ & PSNR$\uparrow$  & SSIM$\uparrow$ \\
    \midrule
    \multirow{3}[2]{*}{RICE-I} & Base  & 0.0202 & 1.278 & 33.65 & 0.9755 & 0.0169 & 1.194 & 34.84 & 0.9604 & 0.0122 & 0.993 & 37.67 & 0.982 \\
          & CA    & 0.0149 & 1.162 & 36.04 & 0.9785 & 0.0166 & 1.212 & 34.88 & 0.9604 & 0.0117 & 0.978 & 37.89 & 0.9819 \\
          & AC-Attention  & $\bm{0.0138}$ & $\bm{1.078}$ & $\bm{36.66}$ & $\bm{0.9786}$ & $\bm{0.0165}$ & $\bm{1.176}$ &$ \bm{35.01}$ & $\bm{0.9609}$ & $\bm{0.0112}$ & $\bm{0.955}$ & $\bm{38.15}$ & $\bm{0.9822}$ \\
    \midrule
    \multirow{3}[2]{*}{RICE-II} & Base  & 0.0222 & 1.859 & 32.34 & 0.8799 & 0.0218 & 1.692 & 34.15 & 0.8518 & 0.0174 & 1.385 & 35.76 & 0.9002 \\
          & CA    & 0.0213 & 1.657 & 33.17 & 0.8876 & 0.021 & 1.664 & 34.6  & 0.8567 & 0.0172 & 1.369 & 35.83 & 0.9003 \\
          & AC-Attention   & $\bm{0.0201}$ & $\bm{1.600}$   & $\bm{33.7}$  & $\bm{0.8911}$ & $\bm{0.2093}$ & $\bm{1.629}$ & $\bm{34.62}$ & $\bm{0.8592}$ & $\bm{0.0169}$ & $\bm{1.364}$ & $\bm{35.93}$ & $\bm{0.9004}$ \\
    \bottomrule
    \end{tabular}%
}
\begin{tablenotes}
\footnotesize
\item[*] 
AC-Attention is integrated into three network architectures: DSen2-CR \cite{meraner2020cloud}, encoder-decoder \cite{deng2023context}, and U-shaped Restormer \cite{zamir2022restormer}, resulting in ACA-CRNet, ACA-EDNet, and ACA-UNet. The original versions of these networks serve as the base. In the CA methods, two blocks in each base network are replaced with Contextual Attention (CA) blocks~\cite{2018Generative}. 
In the AC-Attention methods, two CA blocks are replaced with AC-Attention blocks. The $\downarrow$ indicates lower is better, while $\uparrow$ indicates higher is better. Bold numbers indicate the best performance achieved in each metric.
\end{tablenotes}
  \label{tab:ab-effect-3datasets}%
\end{table*}%

\begin{figure}[!t]
	\centering
	\includegraphics[width=3.5in]{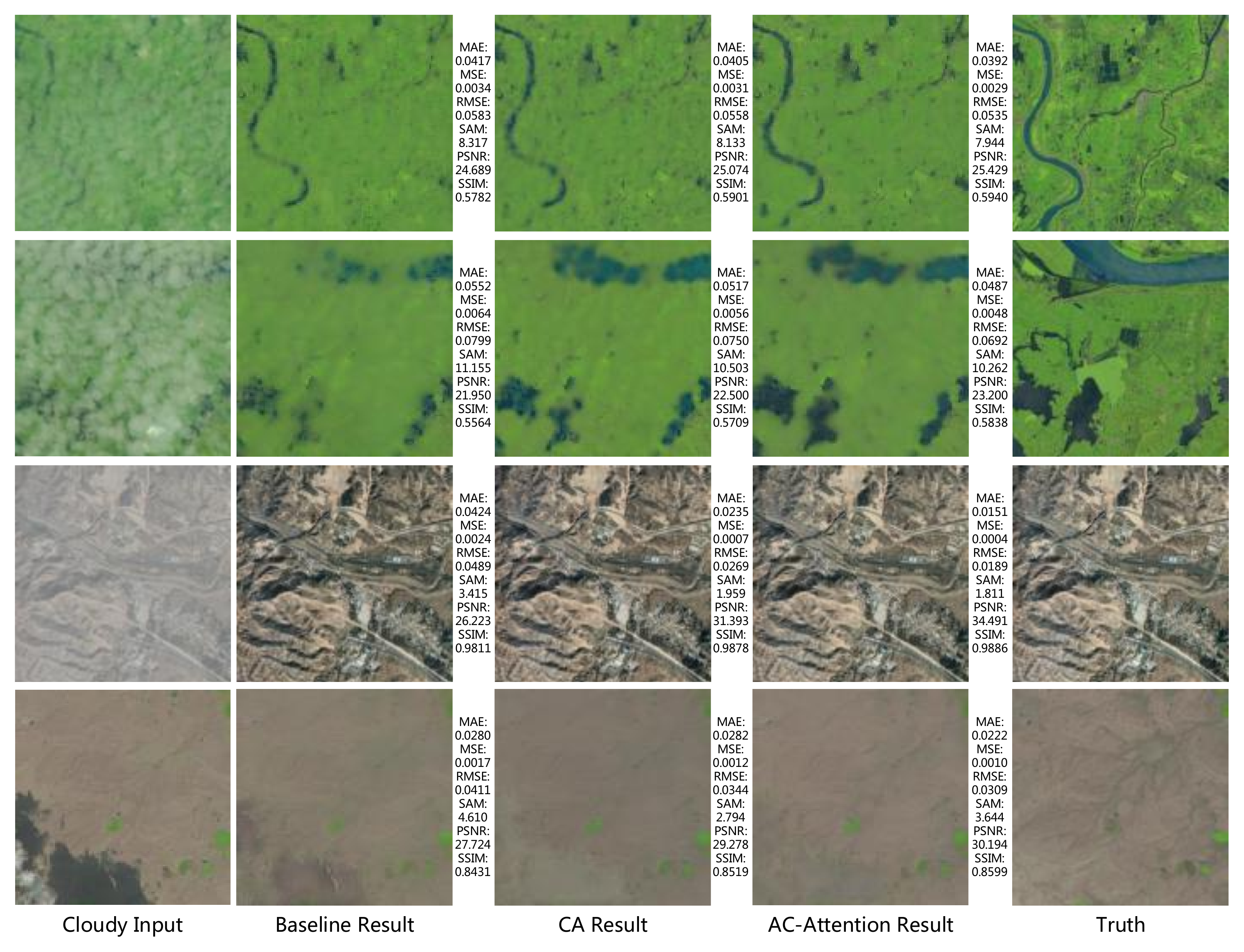}
	\caption{Visualization results of ACA-CRNet demonstrating the effectiveness of our AC-Attention.}
	\label{fig:ab-effect}
\end{figure}

\begin{figure}[!t]
	\centering
	\includegraphics[width=3.5in]{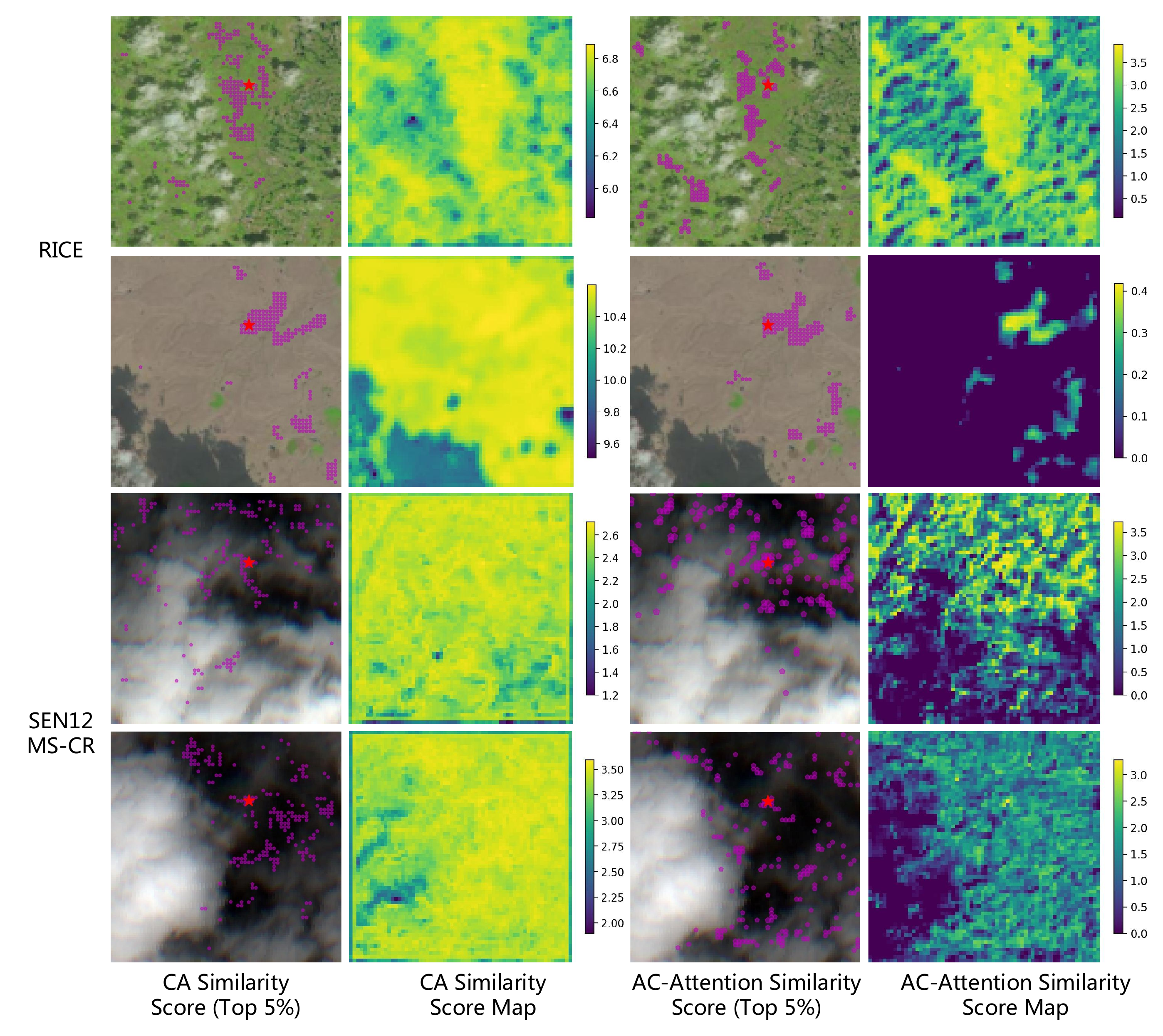}
	\caption{Illustration of Contextual Attention (CA) and our AC-Attention, showing the top 5\% similarity scores and similarity maps. The query is located at coordinates (0.3, 0.6) of the image height and width, marked by a red star, while magenta pentagons indicate points with high similarity scores. The size of each pentagon reflects the score, with larger pentagons representing higher scores.
}
	\label{fig:ab-attentive-score}
\end{figure}

\begin{table}[tbp]
	\centering
	\caption{\chadded{Effectiveness of AC-Attention in Established Methods}  } 
	\resizebox{.999\columnwidth}{!}  {
		\begin{tabular}{cccccc}
			\toprule
			\textbf{Datasets} & \textbf{Methods} & \textbf{MAE}$\downarrow$ & \textbf{SAM}$\downarrow$ & \textbf{PSNR}$\uparrow$  & \textbf{SSIM}$\uparrow$  \\
			\midrule
			\multirow{6}[6]{*}{RICE-I} & SpA GAN~\cite{pan2020cloud} & 0.040  & 2.352  & 28.04  & 0.915  \\
			& \textbf{Ours-SpA GAN} & \textbf{0.025} & \textbf{1.839} & \textbf{31.40 } & \textbf{0.956} \\
			\cmidrule{2-6}          & STGAN~\cite{sarukkai1912cloud} & 0.024  & \textbf{1.701 } & 32.38  & 0.956  \\
			& \textbf{Ours-STGAN} & \textbf{0.023} & 2.223 & \textbf{32.47 } & \textbf{0.956} \\
			\cmidrule{2-6}          & DSen2-CR~\cite{meraner2020cloud} & 0.020  & 1.278  & 33.65  & 0.975  \\
			& \textbf{Ours-DSen2-CR} & \textbf{0.014 } & \textbf{1.078 } & \textbf{36.66 } & \textbf{0.979 } \\
			\midrule
			\multirow{6}[6]{*}{RICE-II} & SpA GAN~\cite{pan2020cloud} & 0.037  & 2.974  & 28.21  & 0.810  \\
			& \textbf{Ours-SpA GAN} & \textbf{0.037} & \textbf{2.518} & \textbf{28.76 } & \textbf{0.835} \\
			\cmidrule{2-6}          & STGAN~\cite{sarukkai1912cloud} & 0.030  & 2.691  & 29.91  & 0.820  \\
			& \textbf{Ours-STGAN} & \textbf{0.028} & \textbf{2.508} & \textbf{30.57 } & \textbf{0.819} \\
			\cmidrule{2-6}          & DSen2-CR~\cite{meraner2020cloud} & 0.022  & 1.859  & 32.34  & 0.880  \\
			& \textbf{Ours-DSen2-CR} & \textbf{0.020 } & \textbf{1.600 } & \textbf{33.70 } & \textbf{0.891 } \\
			\bottomrule
		\end{tabular}%
	}
	\begin{tablenotes}
	\footnotesize
	\item[*] The Ours-* variants modify established methods by replacing the CNN block with our AC-Attention block. The $\downarrow$ indicates lower is better, while $\uparrow$ indicates higher is better. Bold numbers indicate the best performance achieved in each metric.
	\end{tablenotes}
	\label{tab:ab-oth-method}%
\end{table}%

\subsubsection{Effectiveness of Our AC-Attention}\
\par
\thinspace‌ \textbf{Effectiveness of AC-Attention in Established Methods.}
\chadded{We integrated the AC-Attention block as a replacement for conventional CNN blocks within three established methodologies: SpA GAN~\cite{pan2020cloud}, STGAN~\cite{sarukkai1912cloud}, and DSen2-CR~\cite{meraner2020cloud}. Empirical evaluations on the RICE-I and RICE-II datasets demonstrated substantial improvements across all performance metrics, as detailed in Table~\ref{tab:ab-oth-method}. Compared to the original implementations, our approach achieved average enhancements of 20.72\% in MAE, 10.76\% in SAM, 4.93\% in PSNR, and 1.52\% in SSIM. These quantitative results provide compelling evidence that the AC-Attention block significantly enhances cloud removal efficacy across diverse methodological frameworks, highlighting its robustness and broad applicability in remote sensing image enhancement.}

\textbf{Effectiveness of AC-Attention in Several Network Architectures.}
\chreplaced{We assessed the efficacy of our AC-Attention mechanism within three widely utilized network architectures: residual networks, encoder-decoder networks, and U-shaped networks. AC-Attention was implemented in the DSen2-CR residual network~\cite{meraner2020cloud} as ACA-CRNet, in an encoder-decoder network~\cite{deng2023context} as ACA-EDNet, and in a U-shaped Restormer network~\cite{zamir2022restormer} as ACA-UNet, with each network's original structure serving as the baseline. In each implementation, two convolutional blocks were replaced with Contextual Attention (CA) blocks, a recognized patch-based attention method \cite{2018Generative}. We further refined the CA blocks by substituting them with our Residual AC-Attention Blocks, which include an attentive matching operation to minimize irrelevant dependencies and enhance focus, particularly in cloud removal tasks. This modification facilitated a systematic evaluation of AC-Attention's impact on performance across different network architectures.}{We evaluate the effectiveness of our AC-Attention by designing three networks. The first network, ``Base", consists of 2 convolutional layers and 16 Residual Blocks (RBs). The second network, named ``CA", replaces the ninth and thirteenth RB modules of the ``Base" network with residual Contextual Attention (CA) blocks. The CA mechanism is a standard patch-based attention method, and our AC-Attention improved upon it by incorporating an attentive matching operation. This enhancement excludes irrelevant dependencies and enhances the attention mechanism's focus in cloud removal tasks. The third network is our ACA-CRNet, which replaces the CA blocks with Residual Attentive Contextual Attention Blocks (RACABs) in the CA network.}

\chadded{We conducted experiments with these nine networks on the RICE-I and RICE-II datasets, with the quantitative results presented in Table~\ref{tab:ab-effect-3datasets}. Our analysis of four key metrics shows that integrating Contextual Attention (CA) into the basic network configuration enhances model accuracy. Further replacing CA with our AC-Attention mechanism leads to additional improvements in accuracy, underscoring AC-Attention's consistent effectiveness in improving cloud removal performance across various architectures. }

\chreplaced{The visualization results for ACA-CRNet, depicted in Fig.~\ref{fig:ab-effect}, demonstrate that images processed with CA are clearer and more detailed. Moreover, images enhanced with AC-Attention more closely match the ground truth regarding clarity, color, texture, and structure. These findings confirm the adaptability of AC-Attention to various architectures and its efficacy in image cloud removal tasks.}{ We conduct experiments on the RICE-I and RICE-II datasets, the quantitative results are shown in Table~\ref{tab:ab-effect-3datasets} and Fig.~\ref{fig:ab-effect}. Comparing the quantitative and qualitative results, we observe that adding CA to the basic network improved model accuracy in all four evaluation metrics. The visualization results also show clearer and more detailed images. Moreover, replacing CA with our AC-Attention further enhanced model accuracy and brought the visualization results closer to the ground truth image regarding clarity, color, texture, and structure. These findings confirm the effectiveness and  of our proposed AC-Attention for image cloud removal tasks.}

\subsubsection{Attentive Similarity Score Analysis}

We analyze the learnable selection mechanism of AC-Attention by visualizing similarity scores and maps for a specific query in the RICE-II and SEN12MS-CR datasets. We compare AC-Attention with Contextual Attention (CA) and present the results in Fig.~\ref{fig:ab-attentive-score}.

When examining the top 5\% similarity scores, we find that AC-Attention assigns high scores to areas similar to the query, such as grassland and land in the first and second rows. In contrast, CA sometimes assigns high scores to unrelated areas. For instance, in the third row, the query is in a cloud-free region, but CA distributes its similarity score to cloud-covered areas.

When comparing the similarity score maps, we observe that CA assigns high yellow scores across most of the feature maps, with only a few regions showing slightly lower scores. However, these lower scores are still relatively high and above 0 (e.g., around 6.0, 9.0, 1.2, and 2.0). In contrast, AC-Attention concentrates high yellow scores in areas relevant to the query while assigning a similarity score of 0 to unrelated regions, effectively eliminating the influence of irrelevant features.

\chadded{In summary, AC-Attention demonstrates a more focused and accurate selection mechanism compared to CA. By focusing high similarity scores on relevant regions and excluding unrelated areas, AC-Attention enhances the accuracy and quality of cloud removal, reducing the impact of irrelevant features.}

\section{Conclusion}


In this paper, we introduce AC-Attention and integrate it into ACA-CRNet, a residual-style network specifically designed for cloud removal in images. AC-Attention addresses the challenge of irrelevant relationships inherent in traditional attention mechanisms by employing a dynamic and learnable selection process that selectively emphasizes beneficial long-range contextual information. 
ACA-CRNet is strategically designed to focus on cloud-specific features by incorporating residual blocks, AC-Attention, and long residual connections. Quantitative evaluations and visual analyses demonstrate that AC-Attention effectively filters out irrelevant feature relationships, significantly enhancing the network's restoration capabilities. As a result, ACA-CRNet demonstrates remarkable superiority in cloud removal tasks, consistently outperforming state-of-the-art models. Furthermore, we integrate AC-Attention into several existing methods and widely used network architectures, further validating its effectiveness and adaptability.

\chadded{Future research can improve the model's efficiency in several ways. First, we plan to create more efficient attention strategies that capture global features while reducing computational costs, addressing the quadratic complexity of attention mechanisms in high-resolution imagery and resource-limited settings. This may include exploring sparse, linear, or hybrid attention approaches. Additionally, we aim to optimize ACA-CRNet for real-time cloud removal by refining its architecture and implementing faster algorithms to reduce inference time.
}

\section*{Acknowledgments}

This work was jointly supported by the General Program of China Postdoctoral Science Foundation (Grant No. 2020M683490), the Youth Program of Shaanxi Natural Science Foundation (Grant No. 2021JQ-054), the Fundamental Research Funds for the Central Universities (Grant No. 220810004005040309), and Ningbo Science and Technology Bureau Project (Grant No. 2023S167). 
The authors would like to express their gratitude to editors, associated editors, and reviewers for their constructive comments and suggestions.

\bibliographystyle{IEEEtran}
\bibliography{IEEEabrv,IEEEexample}












\newpage

 

\vspace{-33pt}
\begin{IEEEbiography}[{\includegraphics[width=1in,height=1.25in,clip,keepaspectratio]{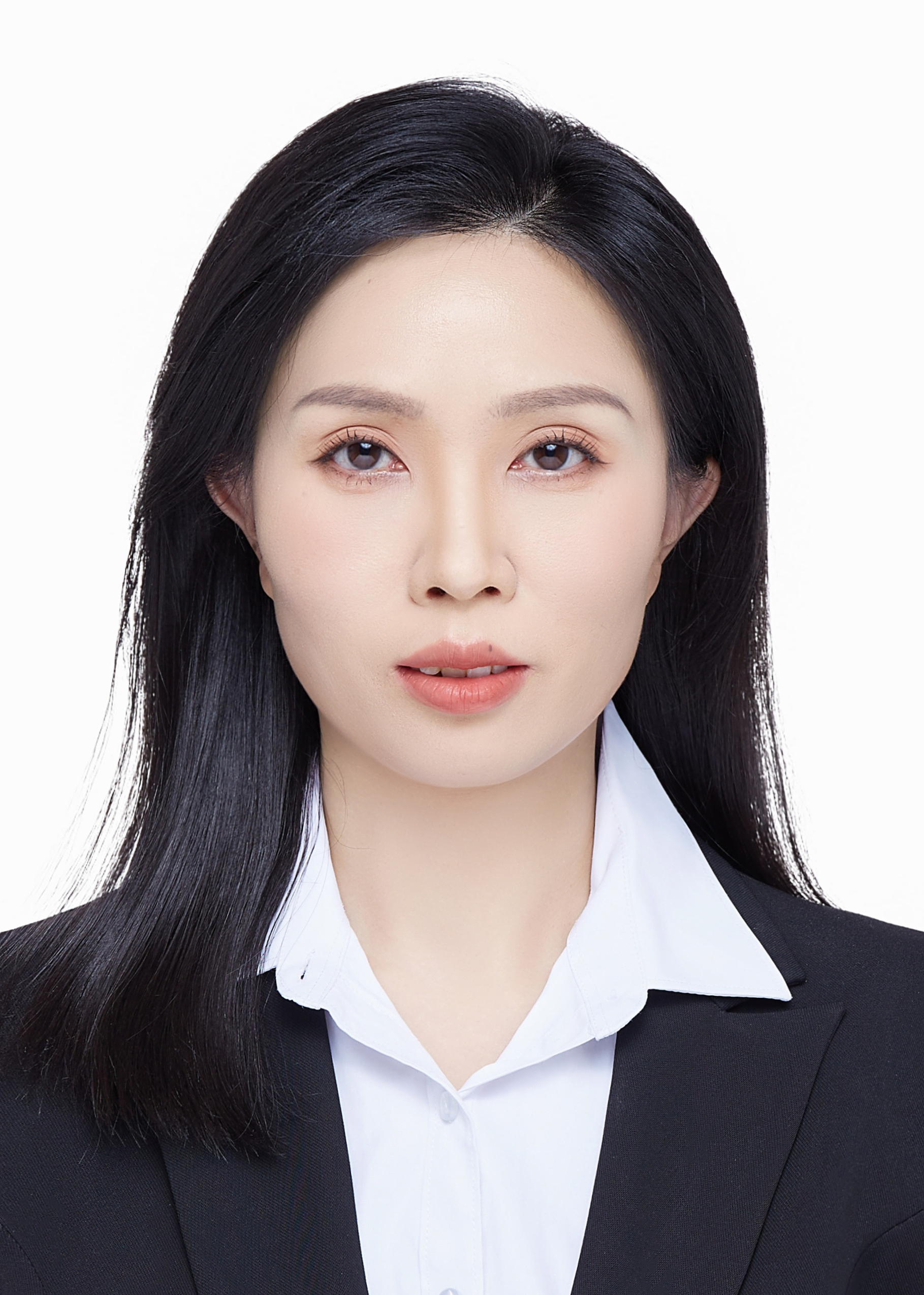}}]{Wenli Huang} received her Ph.D. from Xi'an Jiaotong University, Xi'an, China, in 2023. She is currently a lecturer at Ningbo University of Technology. Her research interests include deep learning, image processing, and computer vision, focusing on network structure optimization, image inpainting, image restoration, etc.

\end{IEEEbiography}
\vspace{-33pt}

\begin{IEEEbiography}[{\includegraphics[width=1.1in,height=1.25in,clip,keepaspectratio]{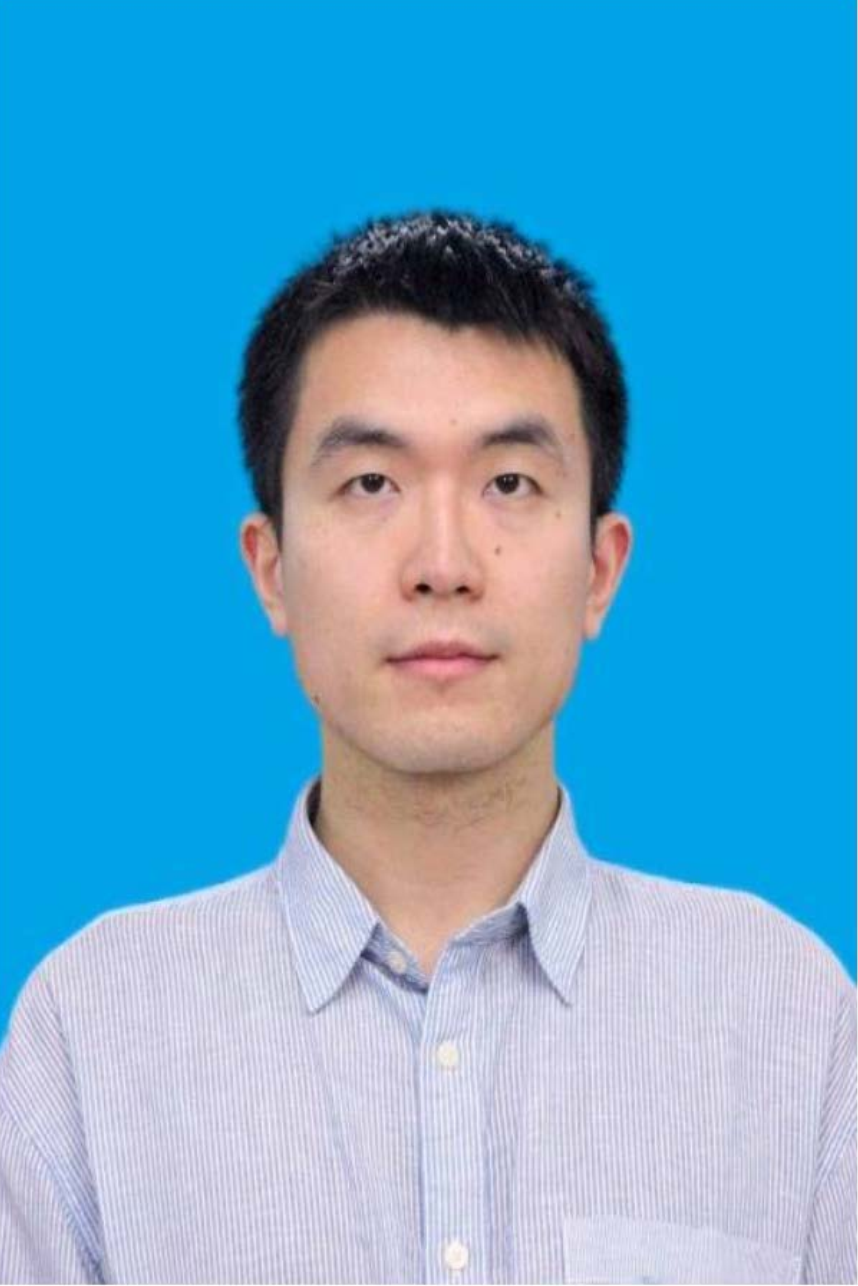}}]{Ye Deng} received his Ph.D. from Xi'an Jiaotong University, Xi'an, China, in 2023. He is currently an Assistant Professor at Southwestern University of Finance and Economics. His research interests include image inpainting, image restoration, and machine learning.

\end{IEEEbiography}
\vspace{-33pt}
\begin{IEEEbiography}[{\includegraphics[width=1in,height=1.25in,clip,keepaspectratio]{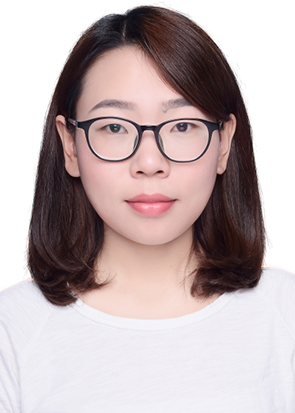}}]{Yang Wu} received the M.E. degree from Northwestern Polytechnical University, Xi’an, China, in 2018. She is currently pursuing a Ph.D. degree in the Institute of Artificial Intelligence and Robotics at Xi’an Jiaotong University. Her research interests include Knowledge graph completion and Graph Representation Learning.

\end{IEEEbiography}
\vspace{-33pt}
\begin{IEEEbiography}[{\includegraphics[width=1in,height=1.25in,clip,keepaspectratio]{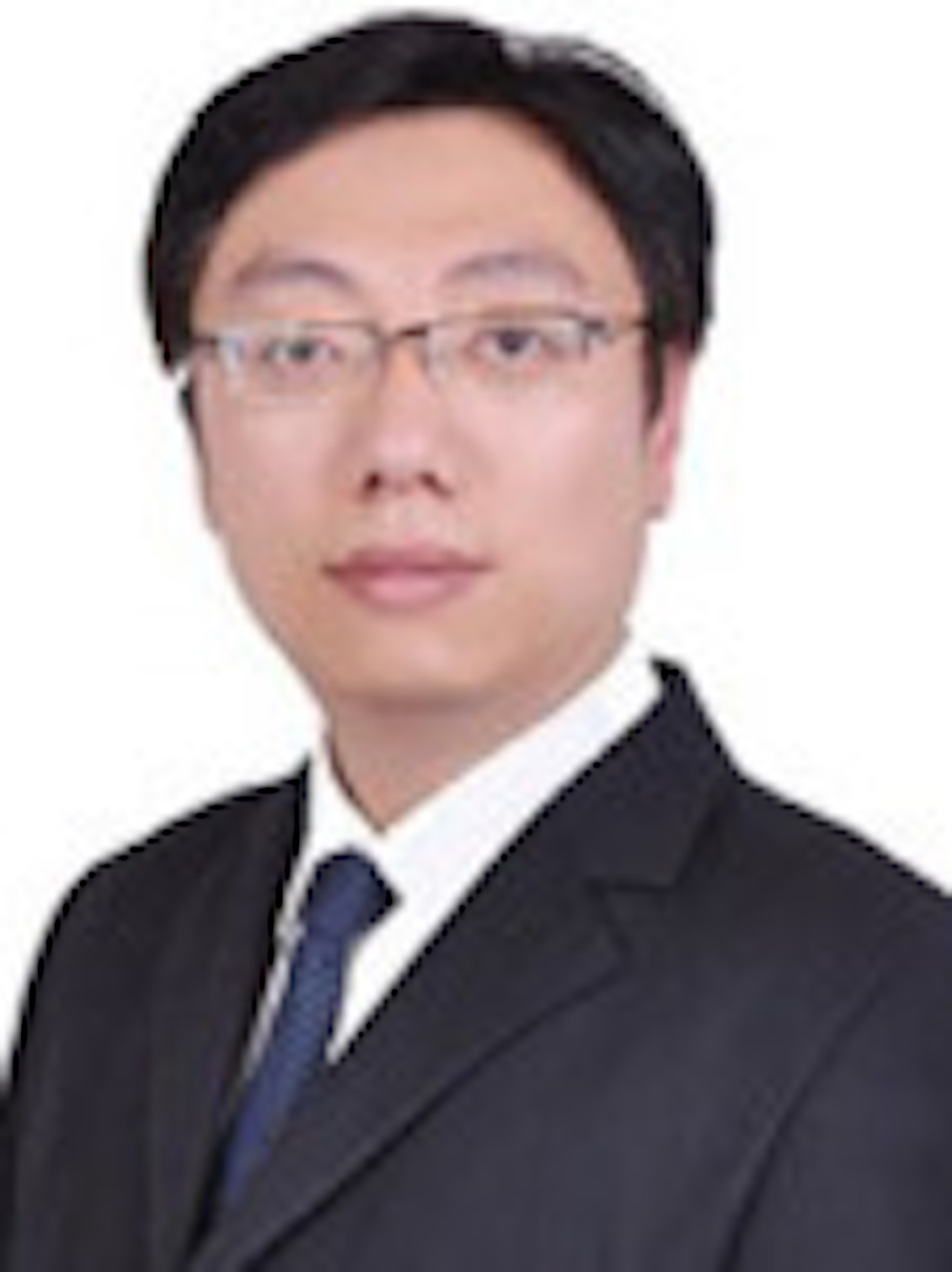}}]{Jinjun Wang} received the B.E. and M.E. degrees from the Huazhong University of Science and Technology, China, in 2000 and 2003, respectively. He received his Ph.D. from Nanyang Technological University, Singapore, in 2006. From 2006 to 2009, he was with NEC Laboratories America, Inc., as a Research Scientist. From 2010 to 2013, he was with Epson Research and Development, Inc., as a Senior Research Scientist. He is currently a Professor at Xi'an Jiaotong University. His research interests include pattern classification, image/video enhancement and editing, content-based image/video annotation and retrieval, semantic event detection, etc.

\end{IEEEbiography}

\vspace{11pt}


\vfill

\end{document}